\documentclass[10pt, conference, compsocconf]{IEEEtran}
\ifCLASSINFOpdf
   \usepackage[pdftex]{graphicx}
   \graphicspath{{../pdf/}{../jpeg/}}
\else
   \usepackage[dvips]{graphicx}
   \graphicspath{{../eps/}}
   \DeclareGraphicsExtensions{.eps}
\fi
\usepackage[cmex10]{amsmath}
\usepackage{algorithm}
\usepackage{algorithmic}
\usepackage[font=footnotesize]{subfig}
\begin{document}
%
\title{Validating Gaussian Process Models \\ with Simulation-Based Calibration}


\author{\IEEEauthorblockN{John Mcleod}
\IEEEauthorblockA{Secondmind\\
Cambridge\\
Email: johnangusmcleod@gmail.com}
\and
\IEEEauthorblockN{Fergus Simpson}
\IEEEauthorblockA{Secondmind\\
Cambridge\\
Email: fergus@secondmind.ai}
}


%


\maketitle

\begin{abstract}
Gaussian process priors are a popular choice for Bayesian analysis of regression problems. However, the implementation of these models can be complex, and ensuring that the implementation is correct can be challenging. In this paper we introduce Gaussian process simulation-based calibration, a procedure for validating the implementation of Gaussian process models and demonstrate the efficacy of this procedure in identifying a bug in existing code. We also present a novel application of this procedure to identify when marginalisation of the model hyperparameters is necessary.
\end{abstract}


%
\IEEEpeerreviewmaketitle

\section{Introduction}
Posterior consistency in Gaussian process (GP) models has been studied analytically (c.f. \cite{CHOI20072975, stuart2018}), however these approaches have focused on the exact GP regression model. This model does not scale to large data sets, so various approximations have been introduced such as the Sparse Variational Gaussian process (SVGP) \cite{SVGP}. These approximations can be sophisticated and may contain subtle errors. As such, we seek an approach for checking the posterior consistency of an arbitrary implementation of a GP model.

Simulation-based calibration (SBC) fits alongside prior and posterior checks and, using simulated data from the prior, inspects the inference procedure which is used to produce the posterior. As such, simulation-based calibration is a diagnostic technique for identifying errors in implementations of Bayesian inference.

The existing SBC literature (c.f. \cite{
talts2020validating, pmlr-v80-yao18a}) assumes a parametric Bayesian model. However the same approach can be applied to a GP model, which is non-parametric. We present Gaussian process simulation-based calibration (GP-SBC), an extension of the SBC algorithm targeting GP models.

\section{Simulation-based calibration for GP models}
\label{sec:GP-SBC}
We refer to \cite{talts2020validating} for the (parametric) SBC algorithm, and \cite{rasmussen2006gaussian} for the Gaussian process regression (GPR) model. Choose a (fixed) set of hyperparameters, a set of training points $X \subset \mathcal{X}$ and a set of test points $X_* \subset \mathcal{X}$ (training and test points may be chosen to reflect a particular data set of interest). Let $\mathbf{f} = f(\mathbf{x})$, then the GP data averaged posterior can be written as:
\begin{equation*}
    p(\mathbf{f}_* | X, X_*) = \int p(\mathbf{f}_* | \tilde{y}, X, X_*) p(\tilde{y} | \tilde{\mathbf{f}}, X) p(\tilde{\mathbf{f}} | X) d\tilde{y} d\tilde{\mathbf{f}} \, .
\end{equation*}

Viewing a GP as a probability distribution over functions provides some intuition for GP-SBC (Algorithm \ref{GP-SBC}), as we can follow the steps of the SBC procedure but replacing the model parameters with a function. The challenge comes from computing the rank statistics of functions, given that a sample from a GP model is the evaluation of the function at an input point. Following \cite{pmlr-v80-yao18a} each output dimension can be considered an independent random variable, and then the images of each of the test points under the prior and posterior functions (in each output dimension) are scalar real values which have a proper ordering. These rank statistics may be plotted as separate histograms for visual inspection or combined into one.


\begin{algorithm}[!t]
\caption{\small{GP-SBC returns $p$ histograms (where $p$ is the output dimension) from the rank statistics of prior function evaluations relative to corresponding posterior function evaluations.}}
\begin{algorithmic}[1]
\small{
    \STATE Given training inputs $X$ and test inputs $X_*$.
    \FOR{$n$ in $N$}
        \STATE Evaluate a function sampled from the prior at $X$ and $X_*$, $\tilde{\mathbf{f}} \sim p(\tilde{\mathbf{f}} | X)$ and $\tilde{\mathbf{f}}_* \sim p(\tilde{\mathbf{f}} | X_*)$
        \STATE Sample a simulated data set, $\tilde{y} \sim p(\tilde{y} | \tilde{\mathbf{f}}, X)$
        \STATE Draw $L$ samples from the posterior at $X_*$
        \FOR{$x_*$ in $X_*$}
            \FOR{$i$ in $\{1, \dots, p\}$}
                \STATE Compute the rank statistic $r$ of the $i$th entry of $\tilde{\mathbf{f}}_*$ relative to the $i$th entries of $\{\mathbf{f}_{*1}, \dots, \mathbf{f}_{*L}\}$
                \STATE Increment the $r$th bin of the $i$th histogram
            \ENDFOR
        \ENDFOR
    \ENDFOR
}
\end{algorithmic}
\label{GP-SBC}
\end{algorithm}

\begin{figure*}[t]
\centering
\subfloat[]{\includegraphics[width=2.5in]{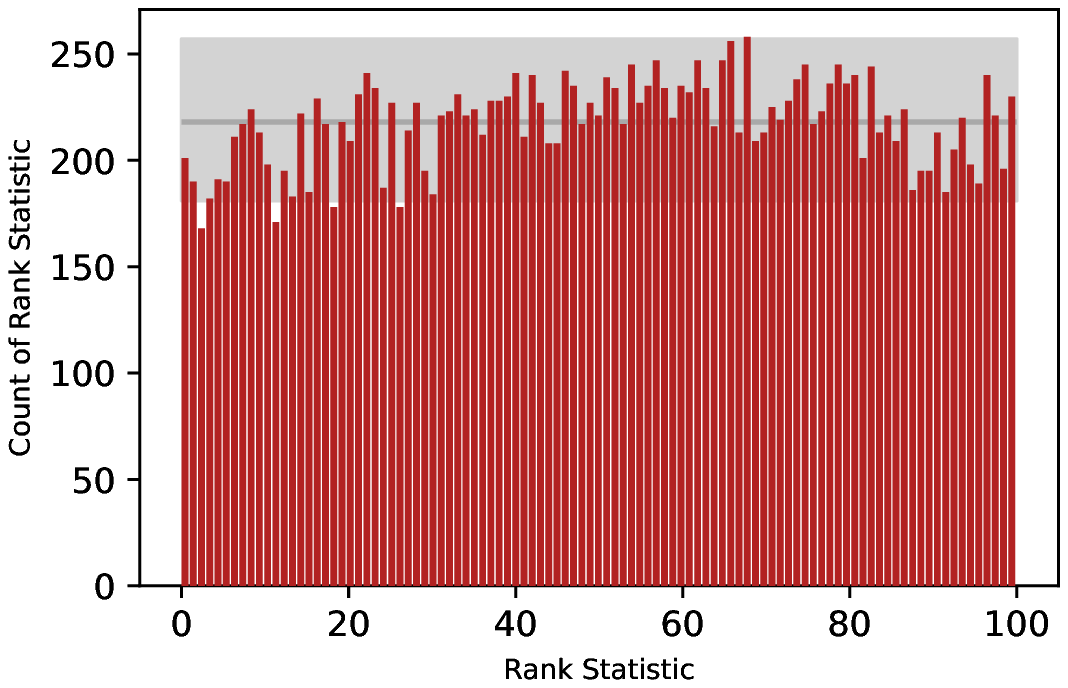}%
\label{fig:MO-SVGP-with-bug}
}
\hfil
\subfloat[]{\includegraphics[width=2.5in]{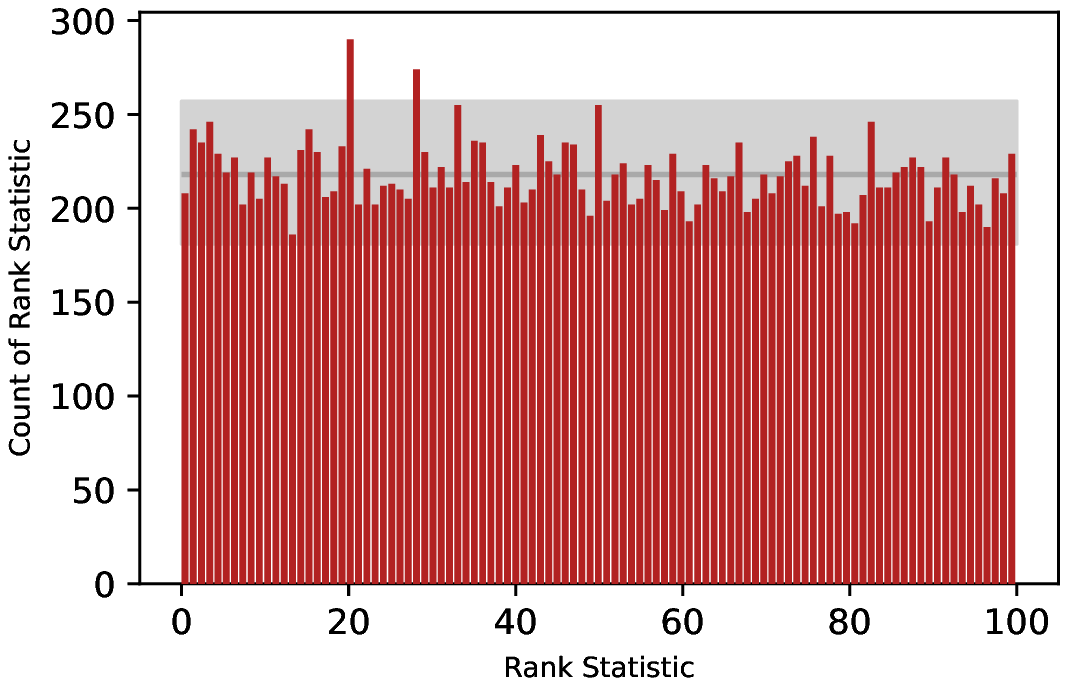}%
\label{fig:MO-SVGP-no-bug}
}
\caption{Rank statistic histogram both before (left panel) and after (right panel) the issue with the multi-output SVGP model was addressed. Following \cite{talts2020validating}, we sample $L = 100$ functions from the posterior and run the algorithm for $N = 1,000$ iterations.}
\label{fig:SVGP-bugfix}
\end{figure*}

\section{Evaluation}
\label{sec:Impl}
To demonstrate the GP-SBC algorithm, we use it to identify a bug in GPflow, one which was recently fixed in version $2.1.5$ (GPflow GitHub pull request 1663). We construct an SVGP model with a linear coregionalization kernel. This model has two independent latent functions, each of which has an squared exponential kernel, and inducing variables. The mixing matrix is the identity and the model has a Gaussian likelihood. We use use the GPflow multi-output framework \cite{vanderwilk2020framework}. We run the GP-SBC procedure on the model both before and after the bug was fixed 
We conclude that there was an issue with the implementation of this model, and that issue has now been addressed. 


\section{Marginalisation of hyperparameters}
\label{sec:marg_hyper}
The GP-SBC algorithm can also be used to identify whether an approximation to the posterior is an inappropriate one. If the approximate posterior is a poor reflection of the true posterior, then this may also trigger a discrepancy in the rank statistic.

Conventionally, when fitting a GP to a data set, the posterior distribution of the model hyperparameters $\theta$ are approximated as a $\delta$-function, i.e. all uncertainties in their values are neglected.  This can be a poor approximation to make, with marginalisation sometimes offering significant gains in performance (c.f. \cite{murray2010slice, filippone2014pseudo, pmlr-v118-lalchand20a, 2020Simpson}). Full marginalisation is rarely used in practice, as it is significantly more computationally expensive.  It would therefore be valuable to be able to identify when marginalisation is necessary, and when we can afford to skip this procedure. The GP-SBC technique can be generalised to address this question.

To begin, a GP model is fitted to the training data, in this case also training the model hyperparameters. We treat this trained model as the prior in the GP-SBC procedure, and use it to generate the simulated data set. Next, replace the model hyperparameters with values sampled from the prior distribution $p(\theta)$, and fit the model to the simulated data.  The remainder of the procedure then runs in accordance with Algorithm \ref{GP-SBC}. If the resulting histogram is inconsistent with a uniform distribution, this indicates that the conventional Type-II maximum likelihood approach has failed: it offers a posterior distribution which is incompatible with the posterior one would have reached had we marginalised over the hyperparameters in a principled fashion. Given that Type-II has a tendency to underestimate the variance of the posterior (c.f. \cite{pmlr-v118-lalchand20a, 2020Simpson}), this failure would be expected to manifest as a `valley' shape in the rank statistic histogram.

\section{Conclusion}

In this paper we introduced the Gaussian process simulation-based calibration algorithm (GP-SBC), an extension of simulation-based calibration to GP models, and demonstrated how the technique can be used to find an inconsistency in the implementation of Bayesian inference. 

Finally we introduced a possible additional application of the GP-SBC algorithm: identifying when marginalisation of model hyperparameters is necessary when constructing the posterior distribution.


\section*{Acknowledgment}

The authors would like to thank S.T. John for help and support with GPflow, and James Hensman for bringing this topic to the attention of the first author.



\bibliographystyle{IEEEtran}
\bibliography{IEEEabrv,bibliography}
%



\end{document}